# On-Line Signature Verification Using Tablet PC

Fernando Alonso-Fernandez, Julian Fierrez-Aguilar, Francisco del-Valle, Javier Ortega-Garcia
Biometrics Research Lab.- ATVS, Escuela Politecnica Superior - Universidad Autonoma de Madrid
Avda. Francisco Tomas y Valiente, 11 - Campus de Cantoblanco - 28049 Madrid, Spain
{fernando.alonso, julian.fierrez, javier.ortega}@uam.es

## Abstract

*On-line signature verification for Tablet PC devices is studied. The on-line signature verification algorithm presented by the authors at the First International Signature Verification Competition (SVC 2004) is adapted to work in Tablet PC environments. An example prototype of securing access and securing document application using this Tablet PC system is also reported. Two different commercial Tablet PCs are evaluated, including information of interest for signature verification systems such as sampling and pressure statistics. Authentication performance experiments are reported considering both random and skilled forgeries by using a new database with over 3000 signatures.*

## 1. Introduction

Automatic signature verification has been an intense research field [1, 2] because of the social and legal acceptance and the widespread use of the written signature as a personal authentication method [3]. Nowadays, there is an increasing use of portable personal devices capable of capturing signature (e.g, Tablet PCs, PDAs, mobile telephones, etc) which is producing a growing demand of person authentication applications based on signature signals.

In this work, we adapt the on-line signature verification system from ATVS to work in Tablet PC environments. The system was evaluated in the First International Signature Verification Competition [4], where was ranked first and second for random and skilled forgeries, respectively.

Additionally, we describe a database captured using two different Tablet PCs. The scope of utility of this database includes the performance assessment in the design of automatic signature-based recognition systems using Tablet PC in several forensic, civil and commercial applications. This new database is used to evaluate the ATVS signature verification system for Tablet PC.

The rest of the paper is organized as follows. The ATVS on-line signature verification algorithm is described in Sect. 2. The application scenario of the prototype developed is described in Sect. 3. The new database is described in Sect. 4. Experimental procedure used to evaluate the prototype and results are described in Sect. 5.

## 2. On-Line Signature Verification Based on HMM

The signature verification system for Tablet PC is based on the recognition algorithm from ATVS presented at the First International Signature Verification Competition (SVC 2004) [4]. This section briefly describes the basics of the recognition algorithm [5]. For further information, we refer the reader to [6], [7] and the references therein. In Fig. 1, the overall system model is depicted.

### 2.1. Feature Extraction

Coordinate trajectories $(x[n], y[n])$, $n = 1, \ldots, N_s$, and pressure signal $p[n]$, $n = 1, \ldots, N_s$, where $N_s$ is the number of samples of the signature, are considered. Signature trajectories are first preprocessed by subtracting the center of mass followed by a rotation alignment based on the average path tangent angle.

An extended set of discrete-time functions are then derived from the preprocessed trajectories. The functions derived consist of a sample by sample estimation of various dynamic properties. As a result, the signature is parameterized as the following set of 7 discrete-time functions $\{x[n], y[n], p[n], \theta[n], v[n], \rho[n], a[n]\}$, $n = 1, \ldots, N_s$, and first order derivatives of all of them, resulting 14 discrete functions ($\theta, v, \rho$ and $a$ stand respectively for path tangent angle, path velocity magnitude, log curvature radius and total acceleration magnitude). A whitening linear transformation is finally applied to each discrete-time function so as to obtain zero mean and unit standard deviation function values.

### 2.2. Similarity Computation

Given the enrolment set of signatures of a client $\mathcal{T}$, parameterized as described in Sect. 2.1, a left-to-right Hidden Markov Model $\lambda^{\mathcal{T}}$ is estimated by using the Baum-Welch iterative algorithm [8], [9]. No transition skips between states are allowed and multivariate Gaussian Mixture density observations are used (2 states and 32 mixtures per state).

On the other hand, given a test signature $S$ (with a duration of $N_s$ samples) and a claimed identity $\mathcal{T}$ modeled as

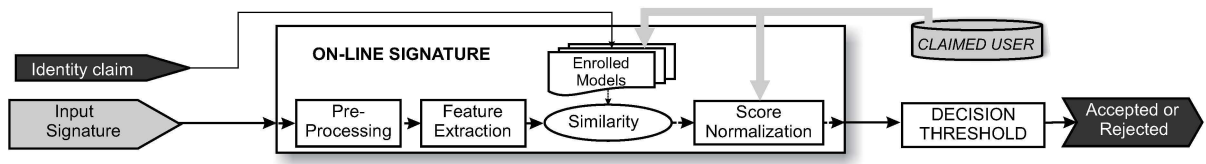

Figure 1. System model for signature person authentication.

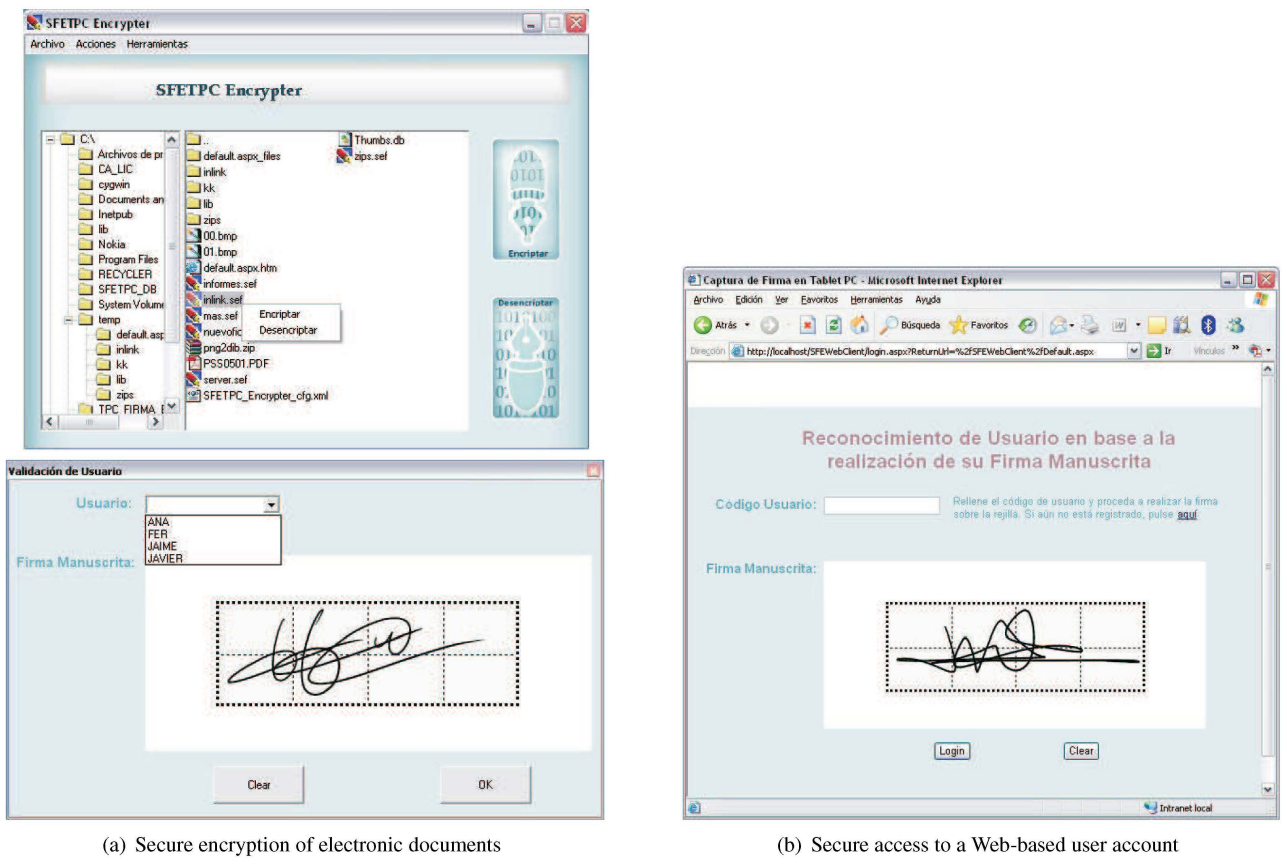

(a) Secure encryption of electronic documents

(b) Secure access to a Web-based user account

Figure 2. Screen captures of the developed prototypes.

$\lambda^{\mathcal{T}}$, the following similarity matching score, $t$, is computed by using the Viterbi algorithm [8]:

$$t = \frac{1}{N_s} \log p\left(S|\lambda^{\mathcal{T}}\right)$$

## 3. Application Scenario

Internet banking, networking, e-Government and other new technologies have been increasingly used in the last years. Ensuring security is an important problem in these environments. Biometric-based solutions to this problem are currently a major research topic [3].

In this framework, we have designed a prototype for $i$) secure access to a Web-based user account and $ii$) secure encryption of electronic documents. Users are enrolled in this prototype by providing 5 signatures in 2 different sessions (3 and 2 signatures in the first and second session respectively).

In Fig. 2, screen captures of the developed prototype are depicted.

## 4. Tablet PC Environment Data

### 4.1. Acquisition

The following Tablet PCs have been used: $i$) Hewlett-Packard TC 1100 with Intel Pentium Mobile 1.1 Ghz processor and 512 Mb RAM, and $ii$) Toshiba Portege M200 with Intel Centrino 1.6 Ghz processor and 256 Mb RAM. Both of them provide the following discrete-time dynamic sequences: position in $x$- and $y$-axis and pressure $p$. Microsoft Windows XP Tablet PC Edition 2005 has been used.

In Table 1, an example of the discrete time sequences provided is shown. Additionally, in Fig. 3, the instantaneous sampling period of two entire signatures is depicted. In Fig. 4(a), the instantaneous sampling period distribution of 5 signatures of different individuals is shown. It can be seen that both Tablet PCs sample at a mean frequency of about 133 Hz but the instantaneous sampling period is not constant. In particular, the sampling frequency for the HP Tablet PC oscillates during the entire signature. To cope with this problem, and reproducing at the same time known signal conditions [4, 6], the position and pressure signals have been downsampled to 100 Hz (constant sampling frequency) using lineal interpolation. The lineal interpolation performed does not introduce relevant distortion in the sequences, since the maximum frequencies of the related biomechanical movements are lower than 20-30 Hz [1].

| HP TC 1100 | | | | Toshiba Portege M200 | | | |
|---|---|---|---|---|---|---|---|
| x | y | p | t (msec) | x | y | p | t (msec) |
| 2262 | 4126 | 19 | 0 | 2454 | 3928 | 24 | 0 |
| 2256 | 4126 | 34 | 2.7682 (+2.7682) | 2441 | 3948 | 54 | 7.442 (+7.4420) |
| 2269 | 4118 | 51 | 14.3258 (+11.5576) | 2430 | 3960 | 85 | 14.9710 (+7.5290) |
| 2267 | 4113 | 66 | 17.8017 (+3.4759) | 2419 | 3964 | 116 | 22.4840 (+7.5130) |
| 2281 | 4092 | 80 | 29.3914 (+11.5897) | 2408 | 3964 | 132 | 30.0130 (+7.5290) |
| 2284 | 4069 | 93 | 32.9374 (+3.5460) | 2408 | 3945 | 146 | 37.5770 (+7.5640) |
| 2305 | 4026 | 104 | 46.2131 (+13.2757) | 2404 | 3920 | 157 | 45.1100 (+7.5330) |
| 2330 | 3971 | 113 | 48.8942 (+2.6811) | 2408 | 3877 | 166 | 54.6920 (+9.5820) |
| 2358 | 3896 | 122 | 61.1508 (+12.2566) | 2413 | 3827 | 174 | 60.1960 (+5.5040) |
| ... | ... | ... | ... | ... | ... | ... | ... |
| Mean sampling period: 7.6353 msec - 130.97 Hz | | | | Mean sampling period: 7.6400 msec - 130.89 Hz | | | |

**Table 1. Discrete time sequences of two signatures provided by the HP TC 1100 and Toshiba Portege M200 Tablet PCs.**

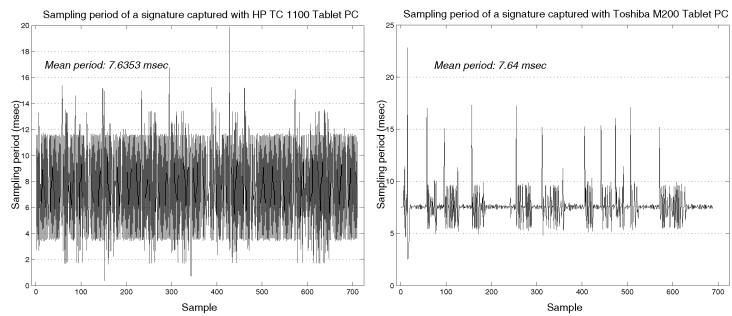

**Figure 3. Instantaneous sampling period of two entire signatures captured with HP TC 1100 and Toshiba Portege M200 Tablet PC.**

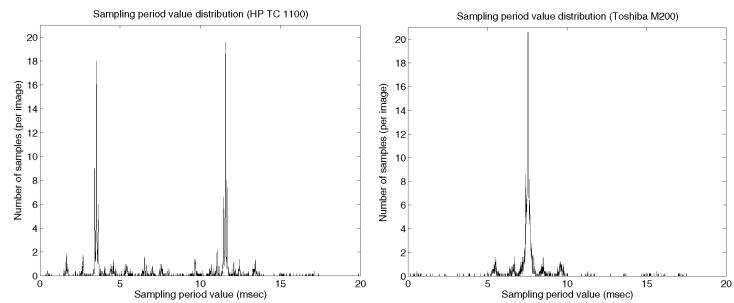

(a) Sampling period distribution

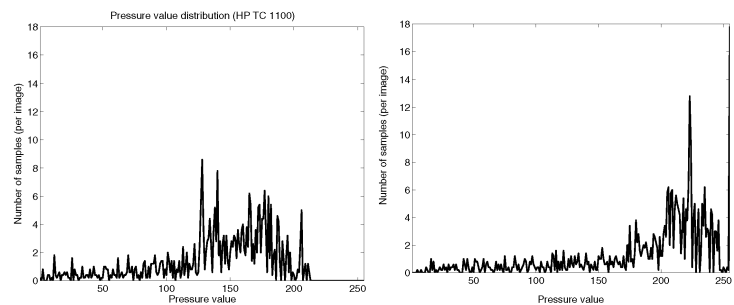

(b) Pressure distribution

**Figure 4. Instantaneous sampling period and pressure distributions of 5 signatures of different individuals captured with HP TC 1100 and Toshiba Portege M200 Tablet PC.**

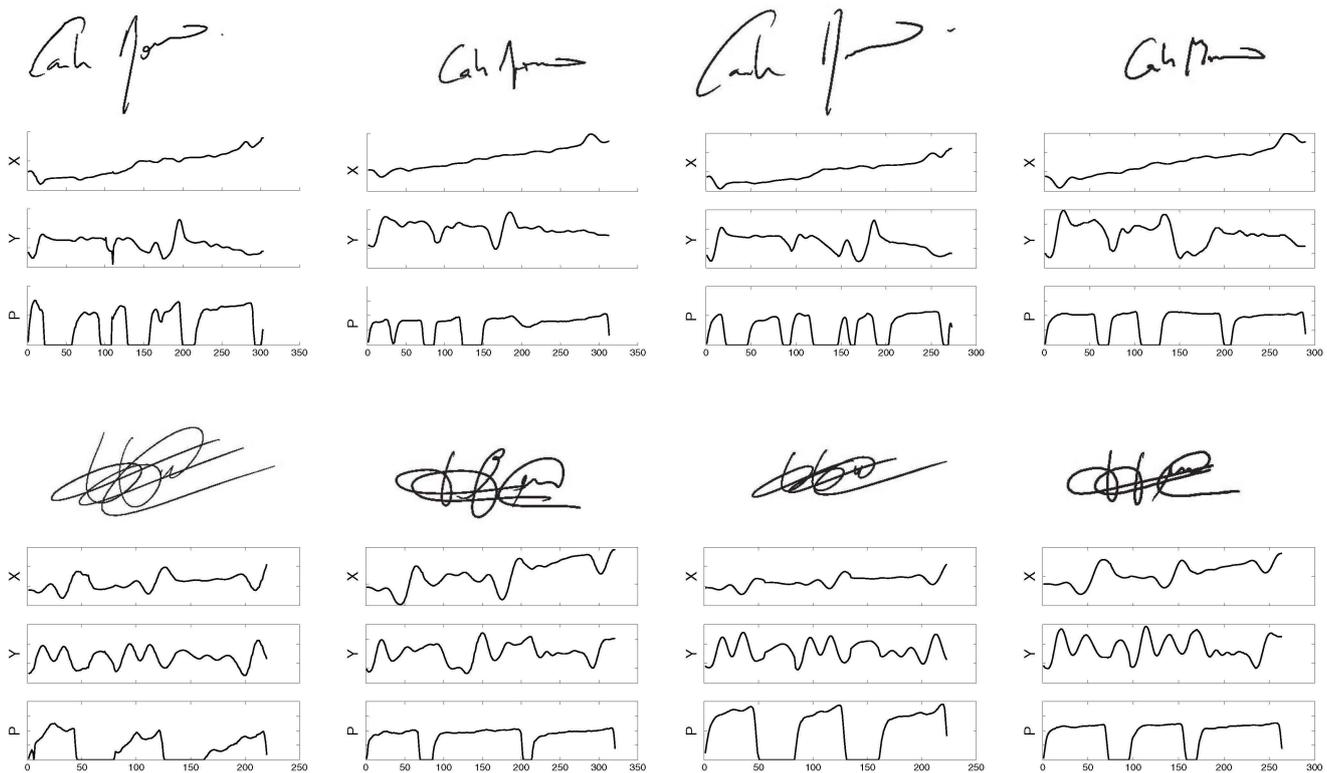

Figure 5. Signature examples captured and interpolated at a constant sampling frequency of 100 Hz. For each row, the left half part (two signatures) corresponds to a subject captured with the HP TC 1100 Tablet PC and the right half part (two signatures) corresponds to the same subject captured with the Toshiba Portege M200 Tablet PC. For a particular subject, the left sample is a client signature and the right one is a skilled forgery. In each case, graph plots below each signature correspond to the on-line information stored in the database.

Regarding to the range of pressure values, our experiments (Fig. 4(b)) show that the two Tablet PCs provide up to 256 values [0-255], although most of the samples are concentrated within a range of approximately 60 pressure values.

### 4.2. ATVS Tablet PC Signature Database

A database of signatures from 53 users has been acquired using the two Tablet PCs. Each user produced 15 genuine signatures in 3 different sessions. For each user, 15 skilled forgeries were also generated by other users. Skilled forgeries were produced by observing both the static image and the dynamics of the signature to be imitated. The dynamics were shown with an animation of the signature using a software viewer. This software also represents the pressure signal as different line thickness in the static signature image. Pen-ups are shown in the animation with a different color in the screen. Imitators were requested to observe the dynamic animation of the signature and to practice the imitation until they were satisfied with it. They were also reminded that the imitation should not be limited to spatial similarity in the shape but should also include temporal similarity.

The acquisition procedure has been performed as follows: user $n$ realizes 5 samples of his/her genuine signature, and then 5 skilled forgeries of a randomly-selected signature of the client $n-1$. Then, again 5 new samples of his/her genuine signature, and then 5 skilled forgeries of a randomly-selected signature of user $n-2$. Then, 5 new samples of his/her genuine signature and finally, 5 skilled forgeries of a randomly-selected signature of user $n-3$. This procedure results in $53 \times 3 \times 5 = 795$ genuine signatures and $53 \times 3 \times 5 = 795$ impostor signatures for each Tablet PC.

Each signature is stored in a separate text file according to the format described in [4]. The first line stores the total number of points in the signature. Each of the following lines corresponds to one sample, which is characterized by the following features: $x$-axis value, $y$-axis value, time instant, button status, azimuth, altitude and pressure $p$. The azimuth and altitude values are set to zero, since the Tablet PCs do not provide this information. The button status feature is set to zero for pen-ups and one for pen-downs.

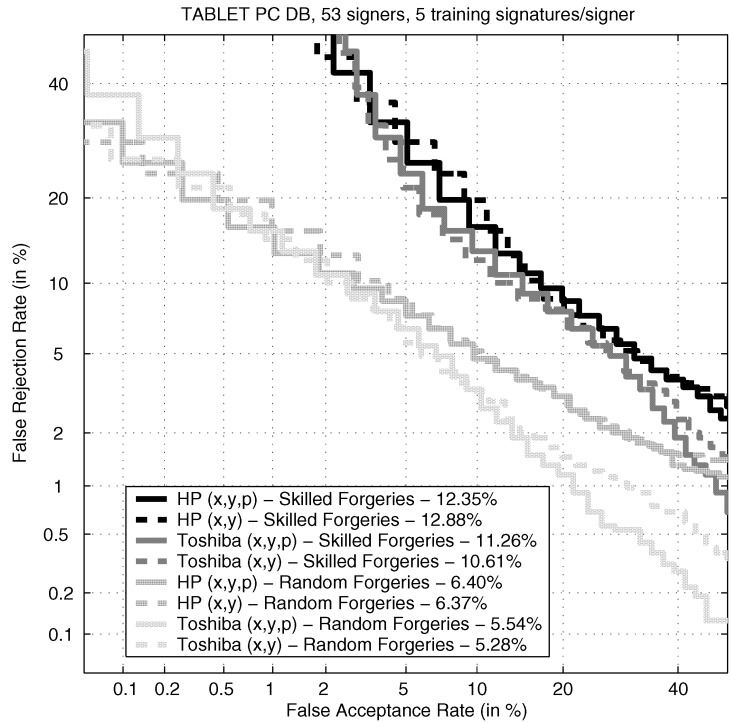

Figure 6. Verification performance for skilled and random forgeries with and without considering pressure information. EER values are also provided.

Some example signatures from this database are shown in Fig. 5.

## 5. Experiments

Multiple signature models are estimated for each user. Each model is estimated by using 3 consecutive genuine signatures from the first session and 2 consecutive genuine signatures from the second session. All the possible combinations are made, thus resulting in 12 different enrolment models per user. With this enrolment scheme we imitate the operational conditions of the application scenario described in Sect. 3. The remaining 5 genuine signatures of the third session are used for testing. For a specific target user, casual impostor test scores are computed by using the skilled forgeries from all the remaining targets. Real impostor test scores are computed by using the 15 skilled forgeries of each target. This results in $12 \times 5 \times 53 = 3180$ genuine user scores, $12 \times 15 \times 53 = 9540$ impostor scores from skilled forgeries and $12 \times 15 \times 52 \times 53 = 496080$ impostor scores from random forgeries for each Tablet PC.

We have evaluated our system with and without considering pressure information. Based on the similarity scores obtained, EER rates and DET curves [10] are obtained. In Fig. 6, verification performance of the Tablet PC signature system is given with and without considering pressure information. In both cases, the Toshiba Tablet results in better performance, both with skilled and random forgeries.

Contrary to the results obtained with the same recognition algorithm on data acquired using high quality pen Tablets [6], considering pressure information does not always result in better performance as compared to not considering pressure information. Considering pressure data leads to improved performance in some operating points with the Toshiba Tablet. The HP Tablet performs approximately the same with and without pressure.

## 6. Conclusions and Future Work

The ATVS Signature Verification System algorithm used in SVC 2004 [4] has been adapted for Tablet PC devices. The resulting system has been evaluated with a database captured using the Hewlett-Packard TC 1100 and Toshiba Portege M200 Tablet PCs. They both provide position in $x$- and $y$-axis and pressure $p$.

Toshiba Tablet has resulted in better performance. This fact may be a result of the HP Tablet sampling frequency oscillation observed. On the other hand, considering pressure information does not result in better performance. This may be because most of the pressure values are concentrated within a small range of approximately 60 pressure values.

Future work includes inter-operability experiments by using both Tablet PCs interchangeably and multi-sensor experiments using both Tablet PCs in combination.

## Acknowledgments

This work has been supported by BBVA, BioSecure NoE and the TIC2003-08382-C05-01 project of the Spanish Ministry of Science and Technology. F. A.-F. and J. F.-A. thank Consejeria de Educacion de la Comunidad de Madrid and Fondo Social Europeo for supporting their PhD studies.